\documentclass[sigconf]{acmart}

\usepackage{graphicx}
\usepackage{subcaption}     
\usepackage{booktabs}       
\usepackage{hyperref}
\usepackage{adjustbox}      
\usepackage{soul} 
\usepackage{todonotes}
\usepackage{xcolor}
\usepackage{hyperref}
\usepackage[shortlabels]{enumitem}

\copyrightyear{2022}
\acmYear{2022}
\setcopyright{rightsretained}
\acmDOI{10.1145/3520304.3528770}
\acmConference[GECCO '22]{2022 Genetic and Evolutionary Computation Conference}{July 9--13, 2022}{Boston, USA}
\acmBooktitle{2022 Genetic and Evolutionary Computation Conference (GECCO '22), July 9--13, 2022, Boston, USA}
\acmISBN{978-1-4503-9268-6/22/07}

\title{EvoJAX: Hardware-Accelerated Neuroevolution}

\author{Yujin Tang}
\email{yujintang@google.com}
\affiliation{%
  \institution{Google Brain}}

\author{Yingtao Tian}
\email{alantian@google.com}
\affiliation{%
  \institution{Google Brain}}

\author{David Ha}
\email{hadavid@google.com}
\affiliation{%
  \institution{Google Brain}}

\hypersetup{
colorlinks=true,linkcolor=blue
}

\definecolor{LightGray}{gray}{0.95}

\setlist[itemize]{leftmargin=*} 

\begin{document}

\begin{abstract}

Evolutionary computation has been shown to be a highly effective method for training neural networks, particularly when employed at scale on CPU clusters. Recent work have also showcased their effectiveness on hardware accelerators, such as GPUs, but so far such demonstrations are tailored for very specific tasks, limiting applicability to other domains. We present EvoJAX, a scalable, general purpose, hardware-accelerated neuroevolution toolkit. Building on top of the JAX library, our toolkit enables neuroevolution algorithms to work with neural networks running in parallel across multiple TPU/GPUs. EvoJAX achieves very high performance by implementing the evolution algorithm, neural network and task all in NumPy, which is compiled just-in-time to run on accelerators. We provide extensible examples of EvoJAX for a wide range of tasks, including supervised learning, reinforcement learning and generative art. Since EvoJAX can find solutions to most of these tasks within minutes on a single accelerator, compared to hours or days when using CPUs, our toolkit can significantly shorten the iteration cycle of evolutionary computation experiments.
\\EvoJAX is available at \textbf{\url{https://github.com/google/evojax}}
\end{abstract}

\maketitle


\section{Introduction}

\begin{figure}[!ht]
\vskip -0.10in
\centering        
\includegraphics[width=0.47\textwidth]{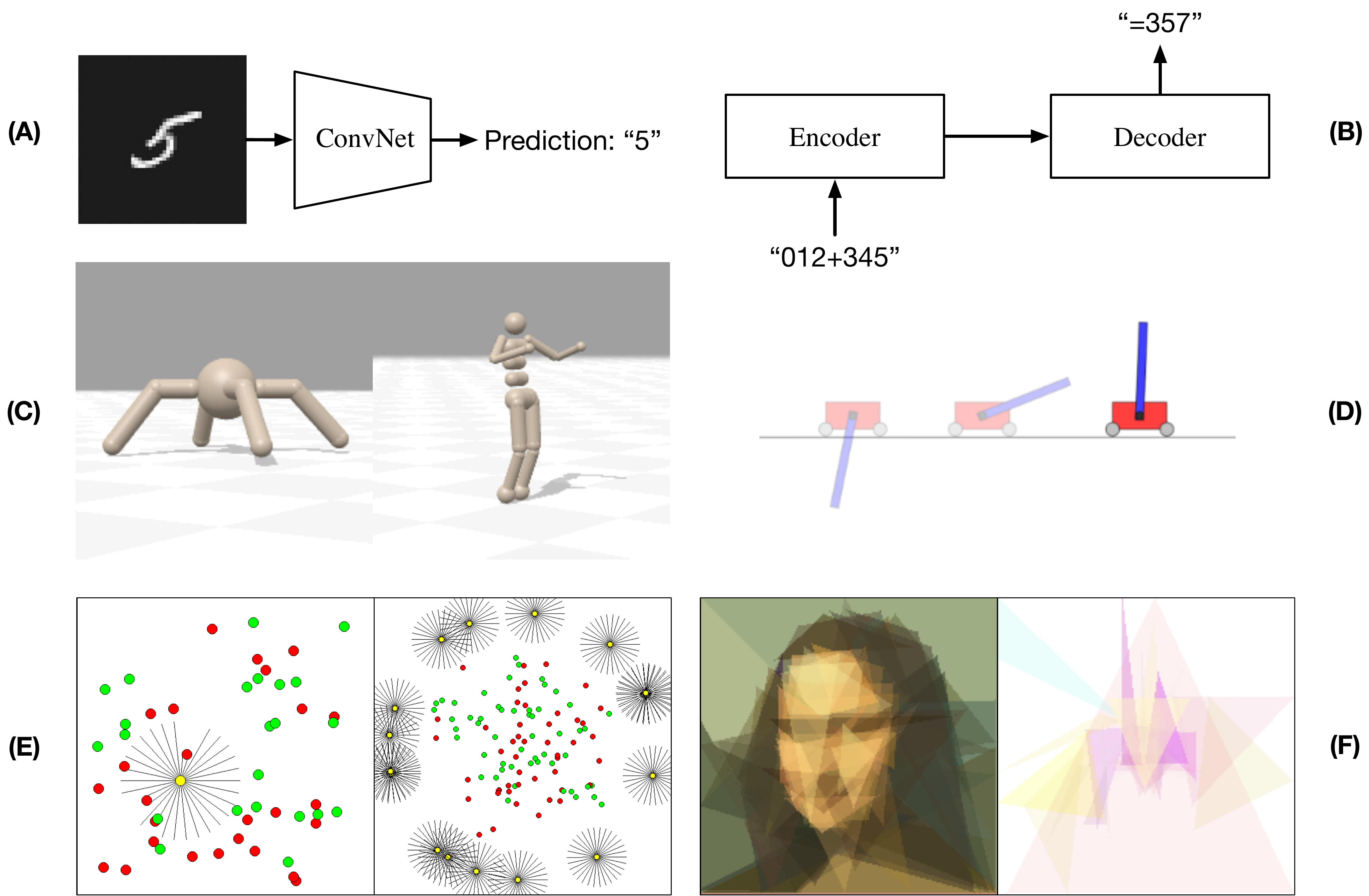}     
\vskip -0.10in
\caption{EvoJAX Examples. (A) MNIST classification. (B) Seq2Seq learning. (C) Robotic control. (D) Cart-pole swing up. (E) Left: WaterWorld wherein the agent (yellow) tries to get food (green) while avoiding poison (red). Right: A version of WaterWorld with multiple agents. (F) Abstract painting with only triangles. Left: Painting a concrete image. Right: Painting the concept ``Walt Disney World''.}
\label{fig:examples_vertical01} 
\vskip -0.20in
\end{figure}

Hardware accelerators have played an important role in advancing the state-of-the-art for deep learning (DL), enabling rapid training of neural networks and shorter research iteration cycles for their development~\cite{hooker2021hardware}. But much of this progress is restricted to systems that rely on gradient descent, a highly effective optimization method when we provide it with a well-defined objective function. But in areas such as artificial life, complex systems, computational biology, and even classical physics~\cite{metz2021gradients}, much of the interesting behaviors we observe take place near the chaotic states, where a system is constantly transitioning between order and disorder. It can be argued that intelligent life and even civilization are all complex systems operating at the \textit{edge of chaos}~\cite{lewin1999complexity,chua2012neurons}. If we wish to study these systems, we need efficient methods to simulate and find solutions in complex systems.

Neural networks are a promising approach for modeling complex systems~\cite{ha2021collective,risi2021selfassemblingAI}, and neuroevolution has made great progress in developing methods for evolving neural networks to solve a wide range of problems. Evolution-based methods have been shown to find state-of-the-art solutions for reinforcement learning (RL)~\cite{salimans2017evolution,such2017deep,jaderberg2017population,tang2020learning,slimevolleygym}. A policy with non-differentiable operations can solve many more tasks than one that is fully differentiable~\cite{wang2019paired,risi2019deep,attentionagent2020,attentionneuron2021}. More importantly, the removal of the requirement of a differentiable policy also liberates the researchers' mind, enabling higher levels of creativity for looking at problems and directions differently from the mainstream. In a sense, enabling researchers to use neural networks beyond gradient-based methods also enables the broader machine learning (ML) research community to explore in a way that is also less ``grad student descent''~\cite{gencoglu2019hark}-based.

However, the progress of hardware-accelerated computational methods for evolution has not kept pace with ML, or even RL. Much of computational evolution is still conducted using CPU clusters, largely ignoring the recent breakthroughs in hardware accelerators such as GPUs/TPUs. Recent work started to demonstrate effectiveness of GPUs for neuroevolution~\cite{such2017deep}, but so far such demonstrations are tailored for specific tasks~\cite{such2017atari}, limiting their applicability to other domains. To enable greater access to hardware accelerators for neuroevolution researchers, we developed EvoJAX, a scalable, general purpose, neuroevolution toolkit. Building on the JAX library~\cite{jax2018github}, our toolkit enables neuroevolution algorithms to work with neural networks running in parallel across multiple TPU/GPUs. EvoJAX achieves very high performance by implementing the evolution algorithm, neural network and task all in NumPy, which is compiled just-in-time to run on accelerators.

In this paper, we describe the design of EvoJAX and show how one can use and extend EvoJAX for neuroevolution research. We showcase several extensible examples of EvoJAX for a wide range of tasks, including supervised learning (image classification, seq-to-seq), RL (cart-pole swing-up~\cite{learningtopredict2019}, Brax locomotion~\cite{brax2021github}, multi-agent water world), and generative art (image approximation with shapes, CLIP-guided abstract art~\cite{tian2021modern}). We show that EvoJAX can find solutions to most of these tasks within minutes on GPU/TPUs, compared to hours or days when using CPUs. We believe our toolkit can significantly shorten the experimental iteration cycle for researchers working with evolutionary computation. We have also created several tutorials and notebooks as part of this open-source project to make adapting EvoJAX for novel use cases straightforward.

\begin{figure}[!t]
\centering

\centering        
\includegraphics[width=0.4775\textwidth]{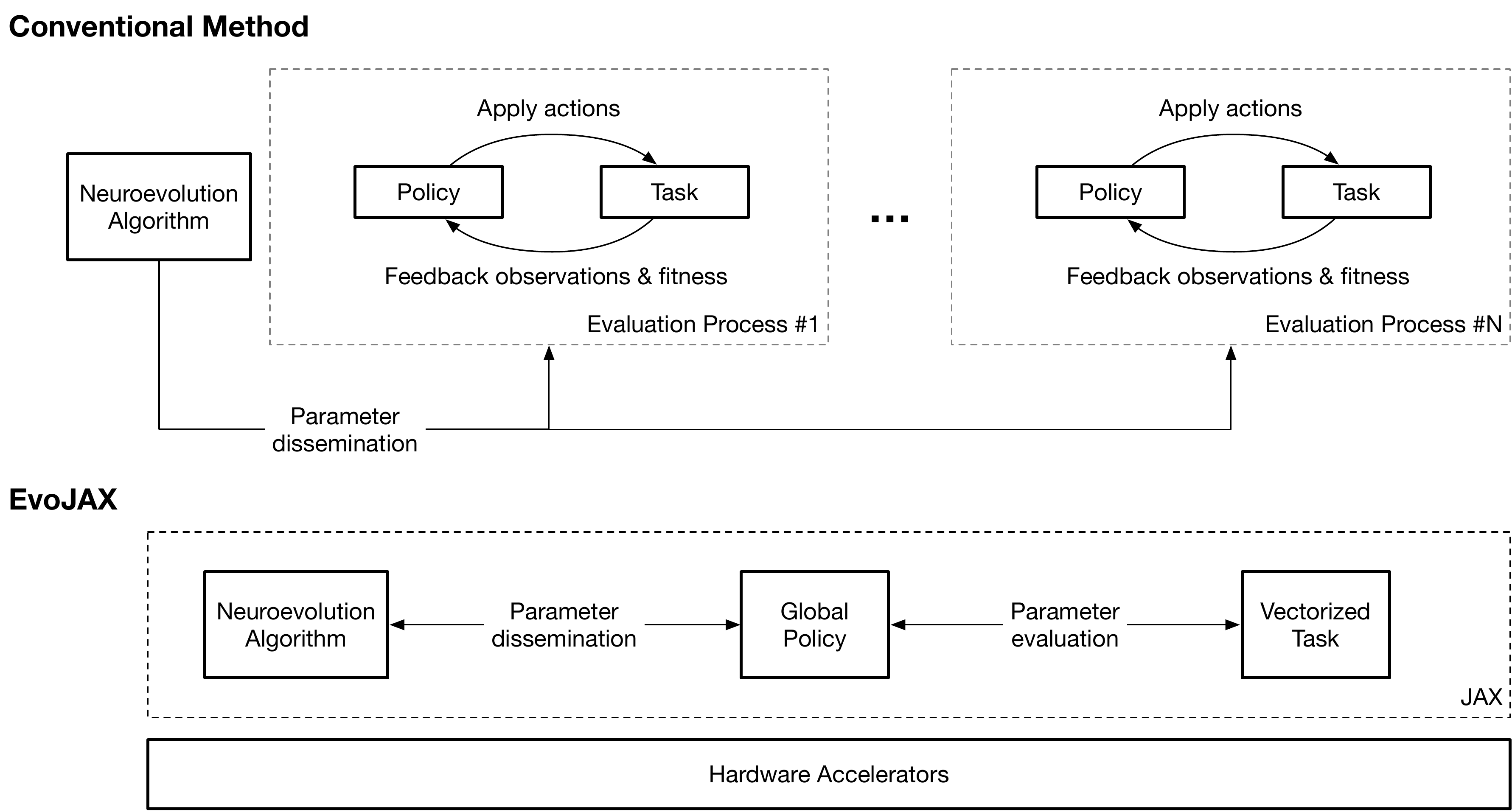}    
\vskip -0.1in
\caption{Architectural Overview of EvoJAX.}
\vskip -0.2in
\label{fig:sys_arch} 
\end{figure}

\section{System Design}

EvoJAX aims to improve the neuroevolution training efficiency by implementing the entire pipeline in modern ML frameworks that support hardware acceleration. We choose JAX\cite{jax2018github} in our current implementation due to its wide variety of hardware support and its matured features of auto-vectorization, device-parallelism, just-in-time compilation, etc. As we will see in Section~\ref{sec:extending_evojax}, as long as the component interfaces are properly implemented, EvoJAX also allows user extensions with other frameworks.

Figure~\ref{fig:sys_arch} gives an overview of how EvoJAX works. There are three major components -- the neuroevolution algorithm, the policy and the task. Although these components are common in conventional neuroevolution implementations, we highlight the key differences that make EvoJAX much more efficient:

\textbf{Modern ML Optimizers}\;\;Researchers and practitioners in the field of DL have been focusing on inventing optimization algorithms~\cite{ruder2016overview} and techniques~\cite{you2019does,keskar2016large,van2017l2} that are both fast and effective. Although these techniques were tailored for gradient-based optimizations, they can be directly applied to gradient estimation-based evolutionary algorithms~\cite{sehnke2010parameter,mania2018simple} too. By leveraging JAX-based libraries~\cite{jax2018github,flax2020github,optax2020github}, EvoJAX not only achieves significant speed-up but also provides the users with the tools and the interfaces to develop their own implementations in a mature framework.

\textbf{Global Policy}\;\;In conventional neuroevolution implementations, it is a common practice to spawn multiple processes for parameters evaluation. To achieve hardware acceleration, the implementation adopts one of the DL frameworks and then each of the evaluation processes maintains a separate computational graph for the same policy. Unfortunately, most DL frameworks are not designed for multi-process training scenarios and often cause difficulties. Moreover, when these processes are run on the same accelerator, maintaining identical copies of the computational graph is a waste of resource. Conforming to the ``Single-Program, Multiple-Data'' (SPMD) model~\cite{darema2001spmd}, EvoJAX solves this by building a global policy and treat both the task observations and the policy parameters as data for the computational graph. This global policy design is easy to implement as it is consistent with DL frameworks, and in the experiments we observe high data-throughput.

\textbf{Vectorized Tasks}\;\;Same as the policies, conventional methods also create copies of the tasks in the spawned processes for independent parameters evaluations. To be compliant with EvoJAX's global policy design, we propose to group these tasks in a vectorized form. In terms of implementation, this can be achieved by either creating the task in auto-vectorizaton supported frameworks or by creating a task observations collector on top of all the evaluation processes. EvoJAX adopts the first method.

\textbf{Device Parallelism}\;\;Thanks to the device-parallelism support in JAX, EvoJAX is capable of scaling its training procedure almost linearly to the available hardware accelerators. Utilizing EvoJAX's training pipeline, this device parallelism is automatically managed and is transparent to the users. As we will see in Section~\ref{sec:examples}, together with the previously mentioned features, EvoJAX significantly shortens the training time for novel and non-trivial tasks.

EvoJAX defines simple yet functionally complete interfaces for the three components, any implementations that are compliant with the interfaces can be seamlessly integrated (see Section~\ref{sec:extending_evojax}).

Finally, in addition to the mentioned major components, EvoJAX also comes with a trainer and a simulation manager that help orchestrate and manage the training process. They contain detailed implementations of task roll-out seeds generation, efficient training loops, time profiling and logistics operations such as logging, testing and periodic model saving. Convenient as they are, we point out that EvoJAX is a flexible toolkit, where it is possible to use any component independently (e.g., using a custom training loop).

\section{EvoJAX Examples}
\label{sec:examples}

We provide a total of six examples (see Figure~\ref{fig:examples_vertical01}) to showcase the capacity, efficiency and the usage of EvoJAX online in the format of Python scripts and notebooks. The examples are designed to feature different aspects of EvoJAX and are in three categories: Supervised Learning Tasks, Control Tasks and Novel Tasks. As the experimental setups, ``Robotic Control'' was trained with TPUs, ``Concrete and Abstract Painting'' was trained with 8 NVIDIA V100 GPUs, and the rest were trained with 1 NVIDIA V100 GPU.

\textbf{Supervised Learning Tasks}\;\;They provide both the data and the ground-truth labels to train the policy. In EvoJAX, supervised learning tasks are modelled as single-step tasks, the examples in this category are thus isolated from other factors to prove the correctness and efficiency of our algorithms' implementation.

\begin{itemize}
    \item \href{https://github.com/google/evojax/blob/main/examples/train_mnist.py}{\textcolor{blue}{\emph{MNIST Classification}}}. 
        Here, we train a convolutional neural network (ConvNet) with ~10K parameters with EvoJAX. Although MNIST is a solved problem in DL, it is non-trivial for neuroevolution in terms of achieving high test accuracy within a short time (e.g., in minutes). We show that EvoJAX can train the ConvNet to reach $>98\%$ test accuracy within 5 minutes. 
    \item \href{https://github.com/google/evojax/blob/main/examples/train_seq2seq.py}{\textcolor{blue}{\emph{Seq2Seq Learning}}}.
        It has recently been shown that genetic algorithms (GA) can train large models~\cite{risi2019deep}. Here, we show that EvoJAX is also capable of training a large network with hundreds of thousands of parameters. We adopt a seq-to-seq task where the policy is required to output a sequence after observing a query sequence. Concretely, the query is a sequence that represents the addition of two randomly generated integers (e.g., ``012+345='', we pad the numbers with leading 0's so that they have equal lengths) and the result is a sequence representing the answer. Using an LSTM-based seq2seq~\cite{sutskever2014sequence} model, EvoJAX achieves $>99\%$ test accuracy within tens of minutes.
\end{itemize}

While one would obviously use gradient-descent for such tasks in practice, the point is to show that neuroevolution can also solve them to some degree of accuracy within a short amount of time, which will be useful when these models are adapted within a more complicated task where gradient-based approaches may not work.

\textbf{Control Tasks}\;\;The purpose of including control tasks are two-fold: 1) Unlike supervised learning tasks, control tasks in EvoJAX have undetermined number of steps, we thus use these examples to demonstrate the efficiency of our task roll-out loops. 2) We wish to show the speed-up benefit of implementing tasks in JAX and illustrate how to implement one from scratch.

\begin{itemize}
    \item \href{https://github.com/google/evojax/blob/main/examples/notebooks/BraxTasks.ipynb}{\textcolor{blue}{\emph{Robotic Control}}}.
        Brax~\cite{brax2021github} is a differentiable physics engine implemented in JAX that simulates environments made up of rigid bodies, joints, and actuators. We show that it is easy to wrap Brax tasks in EvoJAX, and it takes EvoJAX tens of minutes to solve a robotic locomotion task on Colab TPUs.
        
    \item \href{https://github.com/google/evojax/blob/main/examples/train_cartpole.py}{\textcolor{blue}{\emph{Cart-Pole Swing Up}}}.
        Through this classic control task, we illustrate how a task is implemented from scratch in JAX and integrated into EvoJAX's training pipeline. In our implementation, a user can command the initial states to be randomly sampled from a narrow (easy version) or a wide (hard version) range of possible settings, with the latter being much harder to solve. EvoJAX solves both versions within minutes.
\end{itemize}

\textbf{Novel Tasks}\;\;In this last category, we go beyond simple illustrations and show examples of novel tasks that are more practical and attractive to researchers in the genetic and evolutionary computation area, with the goal of helping them try out ideas in EvoJAX.
    
\begin{itemize}
    \item \href{https://github.com/google/evojax/blob/main/examples/train_waterworld.py}{\textcolor{blue}{\emph{WaterWorld}}}.
        In this task~\cite{waterworld}, an agent tries to get as much food as possible while avoiding poisons. EvoJAX is able to train the agent in tens of minutes. Furthermore, we demonstrate that \href{https://github.com/google/evojax/blob/main/examples/train_waterworld_ma.py}{\textcolor{blue}{multi-agents training}} in EvoJAX is possible. Here, we spawn the entire population in the same task roll-out and directly measure each agent's performance in a multi-agent world. This training scheme automatically generates task complexity beyond human design, and is beneficial for learning policies that can deal with interactions between agents and environmental uncertainties.

    \item \href{https://github.com/google/evojax/blob/main/examples/notebooks/AbstractPainting01.ipynb}{\textcolor{blue}{\emph{Concrete}}} \emph{and} \href{https://github.com/google/evojax/blob/main/examples/notebooks/AbstractPainting02.ipynb}{\textcolor{blue}{\emph{Abstract Painting}}}.
        We reproduce the results from a computational creativity work~\cite{tian2021modern}. The original work, whose implementation requires multiple CPUs and GPUs, could be accelerated on a single GPU efficiently using EvoJAX, which was not possible before. Moreover, with multiple GPUs/TPUs, EvoJAX can further speed up the mentioned work almost linearly. We also show that the modular design of EvoJAX allows its components be used independently -- in this case it is possible to use only the neuroevolution algorithms from EvoJAX while leveraging one's own training loops and environment implantation.
\end{itemize}

We summarize EvoJAX's benefit via these examples.
First of all, EvoJAX brings significant training speed up.
In Table~\ref{table:time_comparison} we show the time costs of training some popular tasks with both a conventional setup and EvoJAX.\footnote{We use the code from \cite{attentionneuron2021} as the baseline. For the Locomotion task, we use PyBullet Ant in the baseline and Brax Ant in EvoJAX. The baseline is trained with 96 CPUs.} On modest hardware accelerators, EvoJAX trains $10\sim20$ times faster which leads to quicker idea iterations. Secondly, the capability of training multi-agents in a complex setting that is beyond human design supplies training environmental richness. And finally, EvoJAX puts the entire pipeline on unified hardware setups and that allows the practitioners to simplify complex hardware arrangements. As an example, for the substantial load of computation in our Abstract Painting example, the baseline needs to use both GPUs and CPUs, while EvoJAX only uses GPUs.

%
\begin{table}[!t]
\caption{Time Comparisons. We report the training time for both methods to achieve widely accepted test scores.}
\vskip -0.1in
\label{table:time_comparison}
\begin{tabular}{lrr}
\hline
                 & Baseline & EvoJAX \\ \hline
MNIST           &  36 min     & 3 min  \\ \hline
Cart-Pole Swing Up (Hard Version)  &  37 min     & 2 min  \\ \hline
Locomotion (Ant)$^1$ &  201 min    & 9 min \\ \hline
\end{tabular}
\vskip -0.2in
\end{table}

\section{Extending EvoJAX}
\label{sec:extending_evojax}

A goal of EvoJAX is to provide researchers with an infrastructure that allows fast idea iterations. With EvoJAX it is possible to devise more effective neuroevolution algorithms, to explore novel policy architectures, and to experiment with new  tasks. EvoJAX has carefully defined interfaces, as long as these interfaces are properly implemented, a user extended module can be integrated into the pipeline seamlessly.

%

\begin{figure}[!h]
\vskip -0.1in
\centering        
\includegraphics[width=0.4775\textwidth]{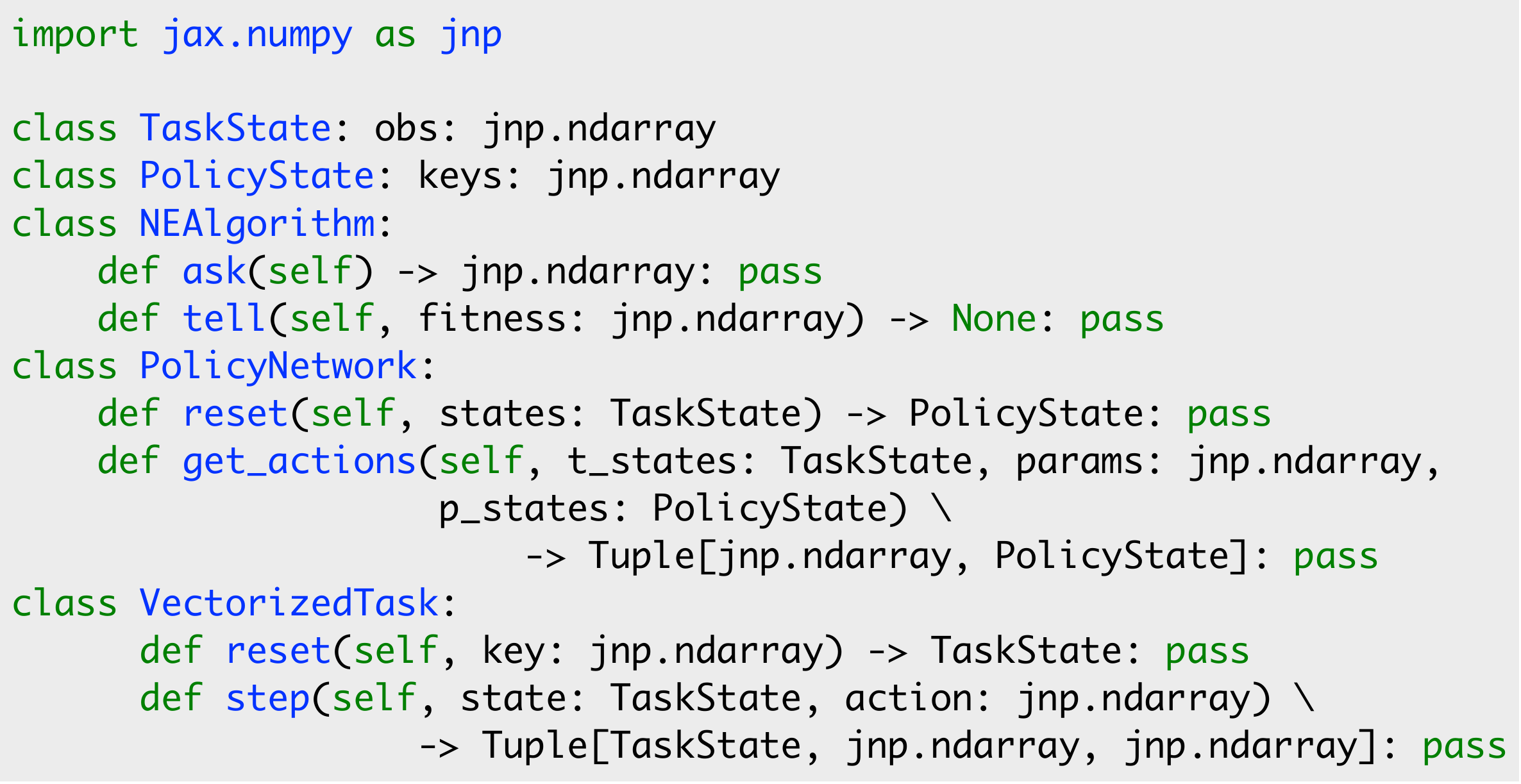}    
\vskip -0.1in
\caption{Major Component Interfaces in EvoJAX.}
\vskip -0.15in
\label{fig:interfaces} 
\end{figure}


\textbf{Devising New Algorithms}\;\;Users interested in inventing new neuroevolution algorithms should implement \textit{NEAlgorithm} in Figure~\ref{fig:interfaces}, which serves as the base class for all neuroevolution algorithms in EvoJAX. Being consistent with most conventional implementations, \textit{NEAlgorithm} adopts the ``ask'' and ``tell'' interfaces, where the former requests the algorithm to generate a population of parameters and the latter reports the parameters evaluation results back to the algorithm for internal states update. Taking on the conventional interfaces for the neuroevolution algorithms not only brings familiarity to the developers and thus reducing the required learning effort, but also allows the practitioners to quickly plug in existing algorithms for sanity checks by writing a simple wrapper.

\textbf{Exploring Novel Policy Architectures}\;\;\textit{PolicyNetwork} in Figure~\ref{fig:interfaces} defines the policy interface, all policies in EvoJAX implement the \textit{get\_actions} method. The method puts no restrictions on what the policy network should be or how it should behave, giving full freedom for neural architecture search (NAS). Because EvoJAX conforms to the SPMD model, \textit{get\_actions} accepts three parameters: the vectorized task states, the population parameters and the policy's internal states. At the beginning of a roll-out, each individual in the population sees identical observations, they will then diverge due to the population's different behaviors. Because JAX requires pure functions, the policy's states (e.g., random seeds, LSTM cell states, etc) are passed to \textit{get\_actions} via a Flax~\cite{flax2020github} dataclass \textit{p\_states}, which is initialized by \textit{PolicyNetwork.reset}. The method returns the actions and the updated policy states. At runtime, calling \textit{get\_actions} is equivalent to passing a batch of data through the model.

\textbf{Experimenting with More Tasks}\;\;In Figure~\ref{fig:interfaces}, \textit{VectorizedTask} forms the base for all EvoJAX tasks. Similar to OpenAI's Gym environments~\cite{brockman2016openai}, the interface defines the \textit{reset} and the \textit{step} methods. Following the pure-function principle of JAX, one major difference between EvoJAX tasks and Gym environments is that EvoJAX's tasks do not keep internal states. Instead, these states are encapsulated in a \textit{TaskState} instance and carried over the roll-out steps. Similar to \textit{PolicyState}, users can inherit \textit{TaskState} and create one's own task specific state to encapsulate arbitrary information besides the environment observations. In most tasks, the initial states are generated via a procedure of randomness. The \textit{reset} method thus accepts \textit{key}'s that act as seeds for the random process.

\section{Limitations and Future Works}

EvoJAX is based on the JAX framework, which is based on the familiar NumPy and is thus friendly to researchers accustomed to such tools. However, practitioners may have to take effort to understand the subtleties of JAX in order to maximize its performance. The time spent on learning the JAX framework may translate to a delayed adoption of EvoJAX, hence much of our focus so far has been on creating examples and tutorials that others can use as templates to build upon. Another limitation of EvoJAX is the compatibility with existing non-parallelizable tasks. Although it is possible to create an observation collector on top of the evaluation processes to mimic the behavior of \textit{VectorizedTask}, the operation involves inter-process communications that becomes a bottleneck, preventing such tasks from the benefit of hardware-acceleration.

In the future, we plan to release more neuroevolution algorithm implementations to EvoJAX in addition to PGPE~\cite{sehnke2010parameter,toklu2020clipup} in the current release. We will add more policies and tasks to both demonstrate a wider variety of examples in order to encourage greater adoption of EvoJAX, with the goal of further enhancing the computation tools available in evolutionary computation research.

\bibliographystyle{ACM-Reference-Format}
\bibliography{reference}

\end{document}